\documentclass[journal]{IEEEtran}
\usepackage{amsmath,amsfonts}
\usepackage{algorithmic}
\usepackage{algorithm}
\usepackage{amsmath}
\usepackage{array}
\usepackage[caption=false,font=normalsize,labelfont=sf,textfont=sf]{subfig}
\usepackage{tabularx}
\usepackage{textcomp}
\usepackage{stfloats}
\usepackage{url}
\usepackage{verbatim}
\usepackage{graphicx}
\usepackage{cite}
\usepackage{hyperref}
\usepackage{multirow}
\usepackage{tabularx}
\usepackage{makecell}
\usepackage{booktabs}
\begin{document}

\title{Flexible Tool Selection through Low-dimensional Attribute Alignment of Vision and Language}


\author{
  \textbf{Guangfu Hao\textsuperscript{1,2,†}}, 
  \textbf{Haojie Wen\textsuperscript{3,†}}, 
  \textbf{Liangxuan Guo\textsuperscript{1,7,†}}, 
  \textbf{Yang Chen\textsuperscript{1}}, 
  \textbf{Yanchao Bi\textsuperscript{4,5,6,*}}, 
  \textbf{Shan Yu\textsuperscript{1,2,7,*}\thanks{†These authors contributed equally to this work. *Corresponding authors: \href{mailto:shan.yu@nlpr.ia.ac.cn}{shan.yu@nlpr.ia.ac.cn}, \href{mailto:ybi@pku.edu.cn}{ybi@pku.edu.cn}}}
\\
  \textsuperscript{1}Laboratory of Brain Atlas and Brain-inspired Intelligence, Institute of Automation\\Chinese Academy of Sciences (CASIA)\\
  \textsuperscript{2}School of Artificial Intelligence, University of Chinese Academy of Sciences (UCAS)\\
  \textsuperscript{3}School of Systems Science, Beijing Normal University\\
  \textsuperscript{4}School of Psychological and Cognitive Sciences \& Beijing Key Laboratory of Behavior and Mental Health, Peking University\\
  \textsuperscript{5} IDG/McGovern Institute for Brain Research, Peking University\\
  \textsuperscript{6} Institute for Artificial Intelligence \& Key Laboratory of Machine Perception (Ministry of Education), Peking University\\
  \textsuperscript{7}School of Future Technology, University of Chinese Academy of Sciences (UCAS)\\
}



\maketitle

\begin{abstract}
Flexible tool selection reflects a complex cognitive ability that distinguishes humans from other species, yet computational models that capture this ability remain underdeveloped. We developed a framework using low-dimensional attribute representations to bridge visual tool perception and linguistic task understanding. We constructed a dataset (ToolNet) containing 115 common tools labeled with 13 carefully designed attributes spanning physical, functional, and psychological properties, paired with natural language scenarios describing tool usage. Visual encoders (ResNet/ViT) extract tool attributes from images while fine-tuned language models (GPT-2, LLaMA, DeepSeek) derive required attributes from task descriptions. Our approach achieves 74\% accuracy in tool selection tasks—significantly outperforming simpler baselines (20\%) and smaller multimodal models (21\%-58\%), while approaching performance of much larger models like GPT-4o (73\%) with substantially fewer parameters. Human evaluation studies validate our framework's alignment with human decision-making patterns, and generalization experiments demonstrate effective performance on novel tool categories. Ablation studies revealed that manipulation-related attributes (graspability, elongation, hand-relatedness) consistently prove most critical across modalities. This work provides a parameter-efficient, interpretable solution that mimics human-like tool cognition, advancing both cognitive science understanding and practical applications in tool selection tasks.
\end{abstract}


\section{Introduction}
\IEEEPARstart{T}{he} ability to flexibly select and use tools represents a remarkable cognitive capability that extends human physical limitations and sets us apart from other species\cite{baber_2003_cognition,johnsonfrey_2003_whats}. While certain non-human animals demonstrate rudimentary tool usage—such as chimpanzees inserting sticks into termite mounds\cite{goodall1964tool}, orangutans using long poles to retrieve fruit\cite{thorpe2009orangutans}, or ants employing leaves to transport food\cite{fellers1976tool}. This fundamental difference in cognitive flexibility represents a key evolutionary advancement that enables humans to exhibit uniquely sophisticated capacity for tool manipulation across diverse contexts.

Beyond basic tool use, humans demonstrate several unique capabilities: designing complex multi-component tools\cite{stout2011stone}, transmitting tool-making knowledge across generations through language\cite{morgan2015experimental}, adapting tools for purposes far removed from their original function\cite{biro2013tool,heersmink2022human}, and creating abstract tools like mathematical symbols and computer algorithms\cite{dehaene2022symbols}. This flexibility allows humans to select appropriate tools for novel situations, repurpose objects for unintended functions, and create new tools to address emerging challenges. Despite its evolutionary significance and centrality to human cognition, the computational and neural mechanisms underlying flexible tool selection remain poorly understood\cite{johnson2004neural,goldenberg_2009_the,vaesen_2012_the,federico2023functional}.

Several neurocognitive studies suggest that the human brain represents tools and their potential uses through abstract attribute spaces rather than rigid categorical classifications \cite{martin1996neural,kellenbach2003actions,huth2012continuous}. When confronted with a novel situation requiring tool use, the brain appears to extract essential functional and physical attributes needed for the task, then matches these requirements against the attributes of available objects. This attribute-based matching process provides a plausible explanation for how humans can generalize tool knowledge to novel situations and identify suitable alternatives when preferred tools are unavailable. 

Despite these insights from cognitive neuroscience, computational models that effectively capture this attribute-based flexible tool selection mechanism remain underdeveloped\cite{osiurak_2018_looking}. Current approaches to modeling tool selection often rely on either direct mapping between task descriptions and tool labels\cite{Saito_2021_How} or multimodal processing that requires extensive computational resources. Additionally, the absence of standardized datasets connecting tool images, usage scenarios, and underlying attributes has impeded progress in this domain.

A key insight from cognitive science research is that humans employ an intermediate level of representation when selecting tools, focusing on functional and physical attributes rather than direct visual-to-task mapping\cite{fischer2021tool}. These attributes serve as a bridge between perception and action, enabling flexible tool use across novel situations. However, formalizing this attribute space and developing computational models that can effectively utilize it remains an open challenge.

In this paper, we propose a novel computational framework that bridges the gap between tool perception and task understanding through a low-dimensional attribute space. Our key contributions include:

\begin{enumerate}

\item We introduce a carefully designed 13-dimensional attribute space that captures both physical properties (elongation, size, hardness) and functional characteristics (graspability, body extension) of tools. We construct four datasets to support attribute-based tool selection research: (1) a tool image-attribute dataset containing 115 common tools with 13 corresponding attribute ratings, (2) a tool scenario-attribute dataset featuring textual descriptions of tool usage scenarios paired with corresponding attribute requirements, (3) a tool matching test set comprising 100 scenario descriptions with 10 candidate tool images each, complemented by human evaluation data from 30 participants, and (4) a novel tool generalization dataset featuring 25 previously unseen tool categories for testing cross-category adaptability.

\begin{figure*}[ht]
    \centering
    \includegraphics[width=1\linewidth]{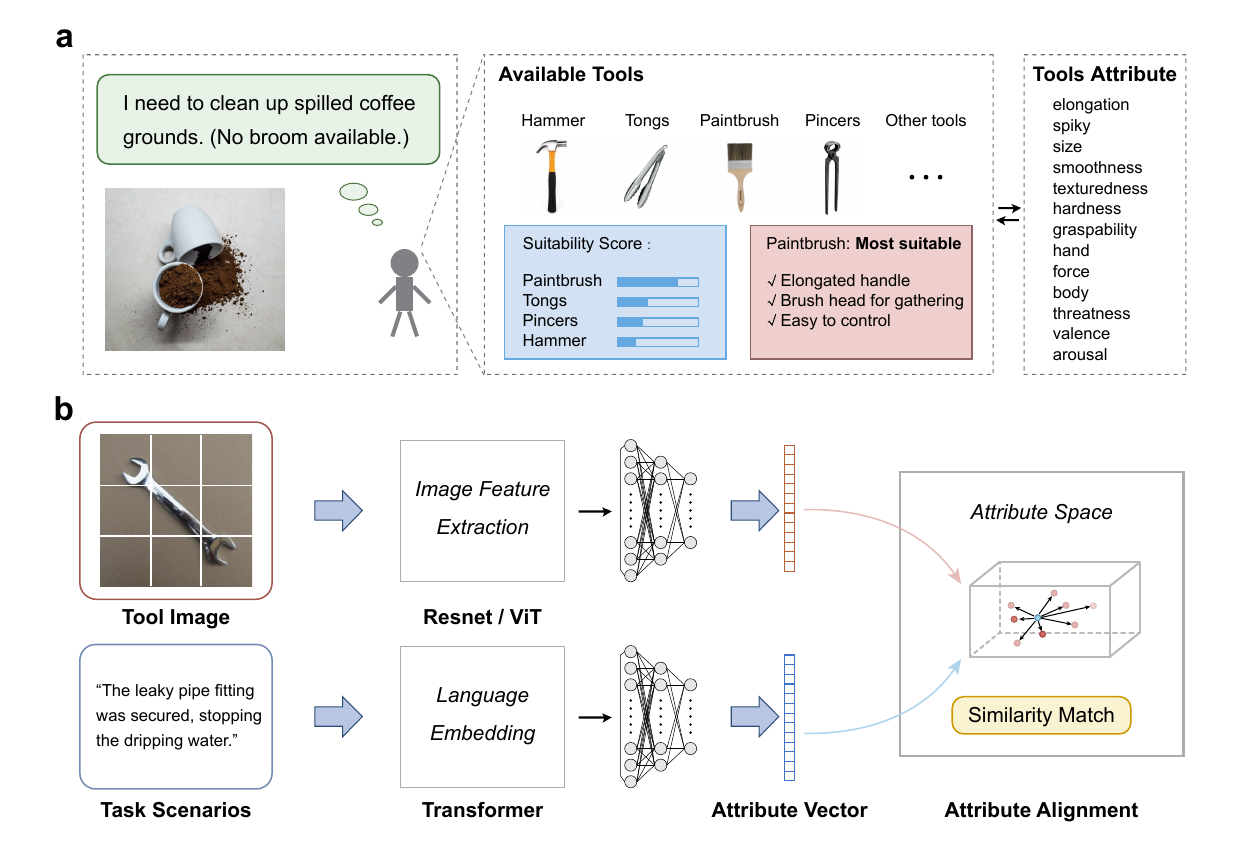}
    \caption{\textbf{Attribute-based flexible tool selection framework.} (a) Example task: When facing a situation like ``I need to clean up spilled coffee grounds (No broom available)'', humans select suitable tools by matching task requirements with tool attributes. Among available tools (Hammer, Tongs, Paintbrush, Pincers, etc.), a paintbrush is selected as most appropriate based on attribute alignment. (b) Our computational framework uses a dual-pathway architecture with a shared 13-dimensional attribute space: the visual pathway extracts attributes from tool images using vision models (ResNet/ViT), while the language pathway derives required attributes from scenario descriptions using LLMs (GPT/Llama/DeepSeek). Tool selection occurs through similarity matching in this shared attribute space.}
    \label{fig:1}
\end{figure*}

\item We propose a low-dimensional attribute space as an interpretable bridge between task requirements and tool representations, enabling flexible tool selection across diverse contexts. We design a dual-pathway attribute alignment method integrating vision and language models. The visual pathway extracts attributes from tool images, while the language pathway derives required attributes from textual scenario descriptions, allowing for cross-modal matching in the shared attribute space.

\item We demonstrate that our attribute-based approach achieves 74\% accuracy in tool selection tasks, substantially outperforming direct tool name matching (20\%) and smaller multimodal large language models (LLMs) (21\%-58\%), while showing competitive performance against much larger multimodal LLMs like GPT-4o\cite{hurst2024gpt} (73\%) and Gemini-2.0-Pro\cite{gemini20pro} (72\%), despite using significantly fewer parameters. Human evaluation studies confirm alignment between our model predictions and human decision-making patterns. Additionally, generalization experiments on novel tool categories demonstrate the framework's ability to extend beyond training data through attribute-based reasoning."

\end{enumerate}

Generalization experiments on novel tool categories further demonstrate that our attribute-based framework can effectively reason about previously unseen tools. This work provides a computationally efficient and cognitively plausible approach to flexible tool selection. By decomposing tool selection into attribute-based representations significantly enhances model performance while requiring substantially fewer parameters than large-scale multimodal LLMs. Through ablation studies, we identify key attributes driving model performance, with functional properties like graspability, elongation, and hand-relatedness proving most critical for accurate tool selection. By demonstrating the efficacy of attribute-based representations in both visual and linguistic domains, our research provides insights into potential mechanisms underlying human tool cognition while offering a practical system that can directly analyze any scenario, process images of available tools, and select the most appropriate tool for the given context.

\section{Related work}

\subsection{Tool Use and Selection in Cognitive Science}

Tool use represents a fundamental cognitive ability that has been extensively studied in both humans and animals. Cognitive neuroscience research has revealed specialized neural mechanisms underlying tool perception and use in humans\cite{peeters2009representation}.  The neural architecture supporting tool use spans multiple brain regions that work in concert to enable the complex cognitive processes underlying this capability.

Neuroimaging studies have identified a specialized tool-processing network primarily in the left hemisphere, including the supramarginal gyrus (SMG), posterior middle temporal gyrus (pMTG), and dorsal premotor cortex (PMd) \cite{gallivan2013decoding}. The left anterior supramarginal gyrus (aSMG) appears uniquely human, specifically devoted to tool use execution and observation \cite{orban2014neural}, highlighting the evolutionary significance of this cognitive ability.

Tool perception integrates multiple cognitive processes beyond mere visual recognition. The parietal lobe facilitates visuo-motor transformations essential for tool manipulation, integrating visual and somatosensory information \cite{maravita2018parietal}. The premotor cortex coordinates the planning and execution of tool-related movements \cite{grafton1997premotor,cabrera2020neural}, while the temporal lobe, particularly the pMTG, stores semantic knowledge about tools and their conventional uses \cite{lesourd2021semantic}. Recent research reveals that the occipito-temporal cortex (OTC) maintains distinct representational spaces for tools, with lateral regions encoding both visual and action-related properties, while ventral areas primarily represent visual features \cite{cortinovis2025tool}.

Several key theories have emerged to explain human flexibility in tool use. The technical reasoning hypothesis proposes that humans possess unique abilities to reason about physical object properties through mechanical knowledge, enabling prediction and analogical transfer across situations \cite{mangalam2022psychological}. This perspective emphasizes abstract conceptual knowledge of functional and physical attributes as the bridge between task requirements and tool selection \cite{osiurak2010grasping,osiurak2016tool}. Neuropsychological evidence from patients with brain damage supports this view, showing similar impairments in both familiar and novel tool use tasks \cite{goldenberg1998tool,osiurak2009unusual}.

In contrast, the manipulation-based approach focuses on sensorimotor affordances and stored action representations. The "Two Action Systems Plus (2AS+)" framework integrates these perspectives, suggesting complementary roles for online reasoning about tool properties and stored manipulation knowledge, with distinct neural substrates supporting each process \cite{buxbaum2017learning}. More recent computational perspectives propose that humans build internal models of tools that enable mental simulation of potential uses before physical interaction \cite{allen2020rapid}.

Despite these advances in understanding the neural and cognitive bases of tool use, computational models that effectively capture human flexibility in tool selection remain underdeveloped. Existing cognitive models typically focus on specific aspects like grasp planning or action execution rather than addressing the broader challenge of flexible tool selection across diverse contexts.

\subsection{Visual Question Answering and Multimodal LLMs}

Tool selection across scenarios fundamentally involves cross-modal reasoning—understanding textual task descriptions while visually evaluating potential tools. This process closely relates to Visual Question Answering (VQA), which has evolved significantly in recent years. Early VQA approaches combined Convolutional Neural Network (CNN)-based visual encoders with Recurrent Neural Network (RNN)-based question processors \cite{antol2015vqa,yang2016stacked}, achieving modest performance through direct feature concatenation. A significant advancement came with attention mechanisms, particularly Bottom-Up and Top-Down attention \cite{anderson2018bottom}, which enabled models to focus on task-relevant image regions based on question content.

Transformer architectures subsequently revolutionized multimodal reasoning by enabling more sophisticated vision-language interactions. Models like ViLBERT \cite{lu2019vilbert}and UNITER \cite{chen2020uniter} adapted masked language modeling objectives to vision-language contexts, learning cross-modal correlations through co-attention mechanisms. These models achieved substantial performance gains by processing textual and visual tokens within unified representational spaces. Contrastive learning approaches further refined cross-modal alignment. CLIP \cite{radford2021learning} demonstrated that training visual and textual encoders to maximize agreement between paired images and captions while minimizing similarity to nonmatching pairs enables powerful zero-shot transfer capabilities.

Current state-of-the-art multimodal LLMs like GPT-4o and Gemini integrate these principles at unprecedented scale \cite{qian2024linguistic}. These models process visual and linguistic information within unified transformer architectures trained on massive multimodal datasets, achieving remarkable performance across diverse VQA benchmarks\cite{xu2024lvlm}. However, they typically require billions to trillions of parameters, substantial computational resources\cite{zhang2024scaling}, and operate as black boxes that obscure their internal reasoning mechanisms\cite{bilal2025llms_tist}. Several specialized multimodal approaches have been developed for tool-related tasks, including systems for robotic manipulation \cite{xie2019improvisation} and visual reasoning about tool functions \cite{myers2015affordance}. The META-TOOL framework \cite{huang2024metatool} specifically evaluated LLMs' ability to determine whether and which tools to select from available options, revealing significant gaps in current models' performance across diverse scenarios.

Although effective, these approaches often require massive model sizes (billions to trillions of parameters) and extensive computational resources for training and inference. Additionally, their black-box nature makes it difficult to interpret their decision processes, particularly in specialized domains like tool selection, where specific functional attributes play a crucial role. The tool selection task represents a distinctive VQA challenge in which the system must understand both the functional requirements implied by a scenario description and the physical capabilities of the available tools.

\subsection{Transfer Learning and Task Alignment}

Efficiently aligning pretrained models with specialized downstream tasks has been extensively studied in both computer vision and natural language processing. These alignment techniques are particularly relevant for attribute-based tool selection, where both visual and linguistic models must be adapted to predict the same attribute space.

\begin{figure*}[t]
    \centering
    \includegraphics[width=1\linewidth]{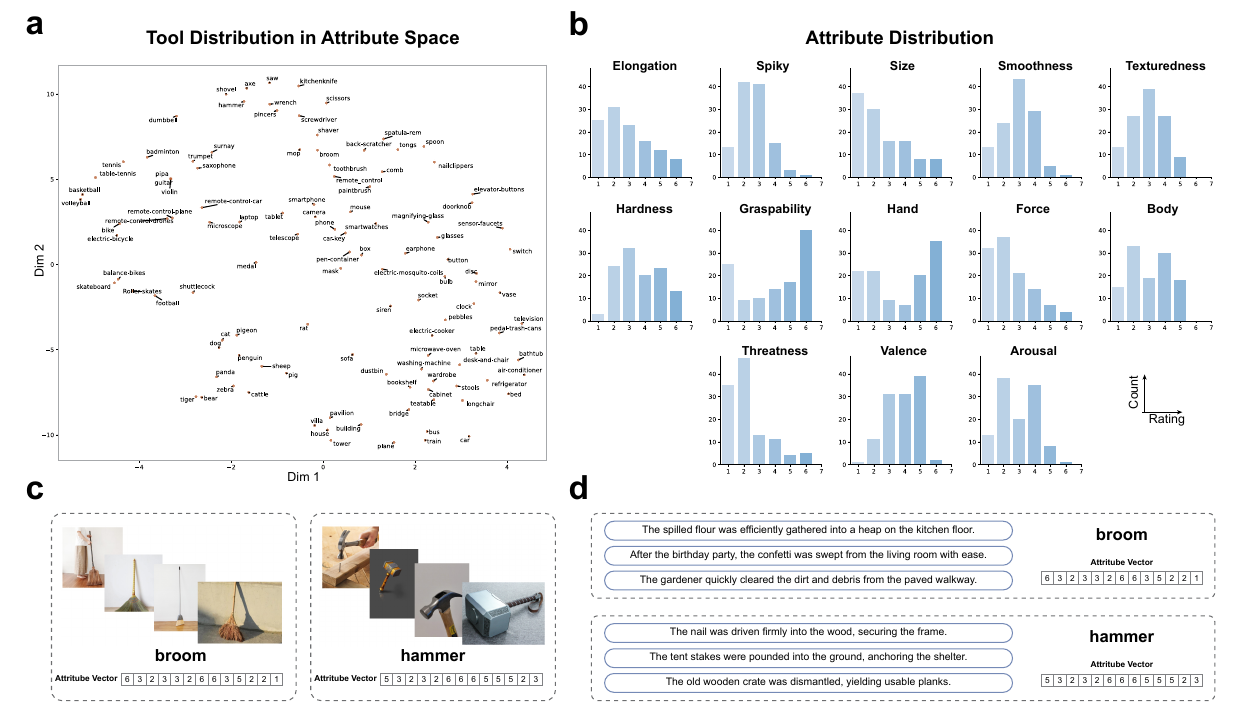}
    \caption{\textbf{Attribute space and the constructed datasets.} (a) Visualization of the 13-dimensional attribute space through dimensionality reduction (PCA), showing well-distributed tool representations that effectively differentiate between tools. (b) Distribution of human ratings across different attributes, demonstrating the variability of attribute values across the tool collection. (c) Sample images from the ToolNet dataset, containing 475 training and 25 testing images per tool across 115 tool categories, with each category sharing the same attribute vector derived from human ratings. (d) Example scenarios from the Task-Description Dataset generated using Gemini 2.0 Flash Experimental LLM, with each scenario associated with a specific tool and inheriting its attribute vector.}
    \label{fig:2}
\end{figure*}

In computer vision, transfer learning has become the dominant paradigm for downstream task adaptation \cite{sharif2014cnn,kornblith2019better}. Convolutional architectures like ResNet \cite{he2016deep} demonstrated remarkable transferability of learned features across diverse visual tasks, while Vision Transformers (ViT)\cite{dosovitskiy2020image} further improved this capability through their attention-based feature extraction. For adapting these pretrained visual backbones to specific downstream objectives, several approaches have proven effective. Linear probing trains only the final classification layer while keeping backbone weights fixed \cite{ninama2024computer}, offering  computational efficiency with minimal risk of catastrophic forgetting \cite{zeng2019continual}. Full fine-tuning adjusts all parameters but typically applies lower learning rates to pretrained layers, enabling more effective adaptation when sufficient task data is available \cite{davila2024comparison}. More parameter-efficient approaches include adapter modules \cite{rebuffi2017learning} that insert small trainable components between frozen layers, and knowledge distillation techniques \cite{beyer2022knowledge} that transfer capabilities from larger teacher models to more efficient student networks.

Parallel to visual model development, language models have evolved from recurrent architectures to transformer-based designs \cite{vaswani2017attention}. Models like GPT \cite{radford2018improving}, BERT \cite{devlin2019bert}, LLaMa \cite{touvron2023llama} and DeepSeek \cite{liu2024deepseek} have demonstrated remarkable capabilities through self-supervised pretraining on massive text corpora, capturing syntactic structure, semantic relationships, and aspects of commonsense knowledge \cite{zhou2020evaluating}.

For language model adaptation, several approaches have proven effective. Traditional fine-tuning adjusts all model parameters on task-specific data \cite{howard2018universal}, consistently delivering strong performance despite computational demands.  Parameter-efficient methods have gained prominence as alternatives: prompt tuning \cite{lester2021power} and prefix tuning \cite{li2021prefix} learn task-specific input tokens while keeping base model parameters frozen, effectively conditioning the model's behavior toward specific outputs.  Low-rank adaptation (LoRA) \cite{hu2022lora} factorizes weight updates into low-rank approximations, dramatically reducing trainable parameters while achieving performance comparable to full fine-tuning.

These developments provide the technical foundation for systems that extract meaningful attributes from both visual tool representations and linguistic usage descriptions—capabilities essential for flexible tool selection. The key challenge in applying these methods to tool selection lies in defining an appropriate attribute space that captures both physical and functional properties relevant to tool use. By adapting pretrained visual and language models to predict the same structured attribute space, we create a bridge between modalities that enables flexible matching of tools to usage scenarios.

\section{Method}
\label{sec:methodology}

\subsection{Dataset Construction}
To enable effective tool selection, we first identified the requirements for bridging visual tool perception and linguistic task understanding. Based on these requirements, we designed a 13-dimensional attribute space that captures the essential physical, functional, and psychological properties of tools. These dimensions were selected based on their theoretical relevance to tool cognition and empirical evidence of their importance in human tool selection processes. The attributes can be broadly categorized into three groups: physical properties (elongation, spikiness, size, smoothness, texturedness, and hardness) that characterize the intrinsic material and structural characteristics of tools; functional properties (graspability, hand involvement, force requirements, and body extension) that describe how the tool interfaces with human users; and psychological properties (threatness, valence, and arousal) that represent the emotional and psychological aspects of tool interaction.

For each attribute dimension, we gathered ratings from 30  participants using a 1-7 scale, with approval from the institutional ethics review board. To ensure consistency, participants were provided with detailed rating guidelines containing dimension definitions, anchor points, and concrete examples. For instance, the elongation attribute was defined as "The degree to which the object is long and slender in shape. A rating of 7 indicates a very elongated object (like a baseball bat), while 1 indicates a non-elongated object (like a disc)." The detailed descriptions and rating criteria for all attributes are shown in Fig. S2.  The final attribute vector for each tool was computed by averaging ratings across all 30 annotators, producing a stable representation of each tool's characteristics in our attribute space. As shown in Fig. \ref{fig:2}(a), dimensionality reduction analysis of the 13-dimensional attribute ratings reveals well-distributed tool representations, indicating that these attributes effectively differentiate between tools. Fig. \ref{fig:2}(b) shows the distribution of ratings across different attributes. While the distributions are not perfectly balanced, this is compensated for by the diversity of tools and multiple images per tool category.

Based on this attribute framework, we constructed three complementary datasets collectively referred to as ToolNet. The first is the Tool Image-Attribute Dataset, a collection of tool images gathered from the internet, with example images shown in Fig. \ref{fig:2}(c). The dataset comprises a training set of 90 images per tool across 115 tool categories (10,350 images total) and a testing set of 10 images per tool (1,150 images total). The tool categories were selected to cover a broad spectrum of everyday tools, spanning domains such as kitchen implements, gardening equipment, workshop tools, and household items. Fig. S1 presents all 115 tool categories with their representative images and names. All images within each tool category share the same attribute vector derived from human ratings.

To enable language-based attribute prediction, we developed Tool Scenario-Attribute Dataset of natural language scenarios describing tool usage contexts, generated using the Gemini-2.0-flash-experimental LLM. The generation process leverages each tool's attribute ratings and attribute descriptions to create natural language descriptions of tool usage scenarios. The detailed prompting strategy used for generating these descriptions is illustrated in Fig. S3. We created three versions of this dataset with varying sizes: the small dataset contains 10 training scenarios and 3 testing scenarios per tool, the medium dataset contains 90 training scenarios and 10 testing scenarios per tool, and the large dataset contains 475 training scenarios and 25 testing scenarios per tool. Example task descriptions are shown in Fig. \ref{fig:2}(d). 

Each scenario is a natural language description of a tool-use situation and inherits the attribute vector of its associated tool. The task descriptions were carefully crafted to maintain linguistic variation while ensuring task relevance. For example, for a broom, scenarios include "The spilled flour was efficiently gathered into a heap on the kitchen floor" and "After the birthday party, the confetti was swept from the living room with ease." This diversity in descriptions helps ensure the robustness of our language encoder in extracting relevant attribute requirements from various phrasings of similar tasks.

To evaluate end-to-end tool selection performance, we constructed Tool Matching Dataset, which contains scenario descriptions paired with multiple candidate tool images for evaluation purposes. This test set includes 100 scenario descriptions, each paired with 1 target tool image and 9 distractor tool images, totaling 1,000 images. The scenario descriptions were extracted from the testing portion of the Tool Scenario-Attribute Dataset, selecting one description for each of the first 100 tool categories. The target and distractor tool images were sourced from the testing portion of the Tool Image-Attribute Dataset.

To establish human performance baselines and evaluate model-human alignment, we conducted a human evaluation study with 30 participants. Each participant evaluated all 100 scenarios from the Tool Matching Dataset, rating the appropriateness of each of the 10 candidate tools for the given usage scenario. Participants used a 7-point Likert scale, where a rating of 7 indicates that the tool is highly suitable for the described scenario, while a rating of 1 indicates complete unsuitability. This evaluation protocol generated 30,000 individual ratings (30 participants × 100 scenarios × 10 tools), providing rich data for analyzing both human decision-making patterns and inter-rater reliability. The study received approval from the institutional ethics review board, and all participants provided informed consent before participation.

To assess our framework's ability to generalize to previously unseen tools, we constructed a New Tool Matching Dataset featuring 25 completely new tool categories not present in the original ToolNet collection. These novel tools, illustrated in Fig. S5, were selected to cover diverse functional categories. For each novel tool, we generated 20 unique usage scenarios using the same protocol employed for the original Tool Scenario-Attribute Dataset, resulting in 500 test scenarios total (25 tools × 20 scenarios). Each test case in the New Tool Matching Dataset consists of a scenario description involving the novel tool paired with 10 candidate tool images: 1 target image of the novel tool and 9 distractor images randomly selected from the original ToolNet collection. This design enables evaluation of whether our attribute-based framework can correctly identify novel tools based on scenario descriptions, even when the visual encoder has never encountered these specific tool categories during training.

This multi-faceted evaluation approach enables comprehensive assessment of our framework across three critical dimensions: computational performance on standard benchmarks, alignment with human decision-making processes, and generalization capability to novel tool categories.

\subsection{Problem Formulation}

We formulate the flexible tool selection task as a cross-modal matching problem in an attribute-mediated space. While traditional approaches might attempt direct mapping between task descriptions and tool categories, our framework operates through an interpretable intermediate attribute representation. Formally, we define the following key components:

\textbf{Attribute Space}: Let $\mathcal{A} \in \mathbb{R}^{13}$ denote our attribute space, where each dimension represents a specific tool property. Each attribute vector $\mathbf{a} \in \mathcal{A}$ consists of elements $a_i \in [1,7]$ corresponding to the rating of the $i$-th attribute. These attributes serve as an interpretable bridge between visual tool representations and linguistic task requirements.

\textbf{Tools}: Let $\mathcal{T}$ be the set of tool images. For each tool category $c \in {1,...,115}$, we have multiple visual instances $\mathbf{t}_c^j \in \mathcal{T}$, where $j$ indexes different images of the same tool category. We define a visual encoder $f_v: \mathcal{T} \rightarrow \mathcal{A}$ that maps a tool image to its attribute representation:
\begin{equation}
\mathbf{a}_t = f_v(\mathbf{t})
\end{equation}

\textbf{Tasks}: Let $\mathcal{D}$ denote the space of natural language task descriptions. The language encoder $f_l: \mathcal{D} \rightarrow \mathcal{A}$ maps a task description to its required attribute representation:
\begin{equation}
\mathbf{a}_d = f_l(\mathbf{d})
\end{equation}

\textbf{Similarity Metrics:} To quantify the compatibility between a task description and a candidate tool, we define a similarity function $s: \mathcal{A} \times \mathcal{A} \rightarrow \mathbb{R}$ that measures the correspondence between attribute vectors. We investigate two primary similarity metrics:

\begin{equation}
s_{cos}(\mathbf{a}_d, \mathbf{a}_t) = \frac{\mathbf{a}_d \cdot \mathbf{a}_t}{||\mathbf{a}_d|| \cdot ||\mathbf{a}_t||}
\end{equation}

\begin{equation}
s_{euc}(\mathbf{a}_d, \mathbf{a}_t) = -||\mathbf{a}_d - \mathbf{a}_t||_2
\end{equation}

where $s_{cos}$ represents cosine similarity and $s_{euc}$ represents negative Euclidean distance.

\textbf{Problem Definition}: Given a task description $\mathbf{d} \in \mathcal{D}$ and a set of candidate tool images ${\mathbf{t}_1,...,\mathbf{t}_n} \subset \mathcal{T}$, the objective is to select the most suitable tool by selecting:
\begin{equation}
\mathbf{t}^* = \arg\max_{\mathbf{t}_i} s(f_l(\mathbf{d}), f_v(\mathbf{t}_i))
\end{equation}

Our framework decomposes the challenging cross-modal reasoning task of tool selection into two more tractable sub-problems: (1) learning to extract relevant attributes from visual tool representations, and (2) learning to infer required attributes from linguistic task descriptions. By operating in this shared attribute space, we enable principled comparison between tools and tasks without requiring massive model sizes or end-to-end multimodal training.

\subsection{Correlation Analysis}
To evaluate the consistency between human evaluators and the similarity between model predictions and human judgments, we employ two complementary correlation measures.

For assessing inter-rater reliability among human evaluators, we calculate Pearson correlation coefficients between all pairs of the 30 human participants. Given two raters $i$ and $j$ with rating vectors $\mathbf{r}_i = [r_{i,1}, r_{i,2}, ..., r_{i,n}]$ and $\mathbf{r}_j = [r_{j,1}, r_{j,2}, ..., r_{j,n}]$ across $n$ tool-scenario pairs, the Pearson correlation coefficient is computed as:

\begin{equation}
\rho_{ij} = \frac{\sum_{k=1}^{n}(r_{i,k} - \bar{r}_i)(r_{j,k} - \bar{r}_j)}{\sqrt{\sum_{k=1}^{n}(r_{i,k} - \bar{r}_i)^2}\sqrt{\sum_{k=1}^{n}(r_{j,k} - \bar{r}_j)^2}}
\end{equation}

where $\bar{r}_i$ and $\bar{r}_j$ are the mean ratings for raters $i$ and $j$, respectively. We compute pairwise correlations for all $\binom{30}{2} = 435$ rater pairs and analyze the distribution of correlation coefficients to assess overall rating consistency.

To evaluate the similarity between model predictions and human preferences, we employ Spearman rank correlation, which measures the monotonic relationship between rankings regardless of the specific numerical values. For each scenario, we obtain both human ratings (averaged across all 30 participants) and model-predicted similarity scores for the 10 candidate tools. Let $\mathbf{h} = [h_1, h_2, ..., h_{10}]$ represent the average human ratings and $\mathbf{m} = [m_1, m_2, ..., m_{10}]$ represent the model scores for the 10 tools in a given scenario. The Spearman correlation coefficient is calculated as:

\begin{equation}
\rho_s = 1 - \frac{6\sum_{k=1}^{10}d_k^2}{10(10^2-1)}
\end{equation}

where $d_k = \text{rank}(h_k) - \text{rank}(m_k)$ is the difference between the ranks of tool $k$ in the human ratings versus model predictions. We compute the Spearman correlation for each scenario and report the mean correlation across all scenarios to quantify overall alignment between model and human tool preferences.

\subsection{Vision-Language Model}
Our framework employs a dual-pathway architecture to bridge visual tool perception and language task understanding through a shared attribute space, as illustrated in Fig.\ref{fig:1}(b). Each pathway is specifically designed to extract relevant attribute information from its respective modality while maintaining interpretability and computational efficiency.

\subsubsection{\textbf{Visual Encoder}}
The visual encoder follows a two-stage architecture comprising a pre-trained feature extractor backbone followed by an attribute prediction head, corresponding to the visual pathway in Fig.\ref{fig:1}(b). We experiment with three backbone architectures: ResNet-18, ResNet-50, and Vision Transformer (ViT-B/16), all initialized with ImageNet pre-trained weights. This design enables efficient transfer learning while preserving the model's capacity to extract tool-specific attributes.

The attribute prediction head is implemented as a multi-layer perceptron (MLP) that maps high-dimensional visual features (512/2048/768-dimensional, depending on the backbone) to our 13-dimensional attribute space. The MLP architecture consists of three fully connected layers: input features → 256 → 64 → 13 (attribute dimensions), with layer normalization and ReLU activation functions after each hidden layer.  This transformation allows the model to distill relevant functional and physical properties from complex visual representations.

\subsubsection{\textbf{Language Encoder}}
The language encoder, representing the language pathway in Fig.\ref{fig:1}(b), maps textual task descriptions to the same attribute space used by the visual encoder. The model consists of a pre-trained language model backbone followed by a specialized regression head.  We experiment with three language model architectures of varying capacities: GPT-2, LLaMA-3.2-1.2B, and DeepSeek-R1-1.5B. The architectural specifications of these models are detailed in Table \ref{table1}, showing substantial differences in parameter count, layer depth, model dimensionality, and attention mechanisms.

\begin{table}[h]
\centering
\caption{Architecture specifications and parameter counts for language models used in our framework.}
\begin{tabular}{lccccc}
\hline
Model Name & $n_\text{params}$ & $n_\text{layers}$ & $d_\text{model}$ & $n_\text{heads}$ & $d_\text{head}$ \\
\hline
GPT-2 & 124.4M & 12 & 768 & 12 & 64 \\
LLaMA-3.2 & 1.2B & 16 & 2048 & 32 & 64 \\
DeepSeek-R1 & 1.5B & 28 & 1536 & 12 & 128 \\
\hline
\end{tabular}
\label{table1}
\end{table}

Each language model processes the input task description and generates contextual representations. We utilize the last token representation for attribute prediction, as this token inherently captures the cumulative context from the entire sequence through the next-token prediction objective. This approach leverages the autoregressive nature of these language models, where the final token embedding contains information about the complete task description. The attribute prediction head transforms the language features into the 13-dimensional attribute vector through a multi-layer architecture consisting of fully connected layers with dimensions [256, 128, 64, 13].  This specialized head is trained to extract attribute requirements implied by natural language task descriptions, enabling cross-modal matching with tool images.

\subsubsection{\textbf{Training Strategy}}
Both encoders are trained to minimize the Mean Squared Error (MSE) loss between predicted and ground truth attribute vectors. The visual encoder is trained with Adam optimizer (learning rate 1e-4) and batch size 256, while the language encoder uses a smaller learning rate (5e-5) and batch size 4 to  balance adaptation with preservation of pre-trained knowledge. For the visual pathway, we freeze the pre-trained backbone (ResNet/ViT) and only train the attribute prediction MLP layers, preventing catastrophic forgetting while allowing specialization to attribute prediction. Similarly, for the language pathway, we keep the pre-trained language model components frozen and only update the regression head parameters. Training employs early stopping based on validation performance, with the visual model trained for up to 1,000 epochs and language models for up to 2,000 epochs.

This dual-pathway architecture enables flexible tool selection by mapping both visual and linguistic inputs to a shared, interpretable attribute space, while maintaining the specific processing characteristics required for each modality. When presented with a task description and candidate tool images, the system computes attribute representations for both and selects the tool whose attributes best match the task requirements, similar to the human process illustrated in Fig.\ref{fig:1}(a).

\section{Results}
\label{sec:result}

\begin{figure*}[t]
    \centering
    \includegraphics[width=1\linewidth]{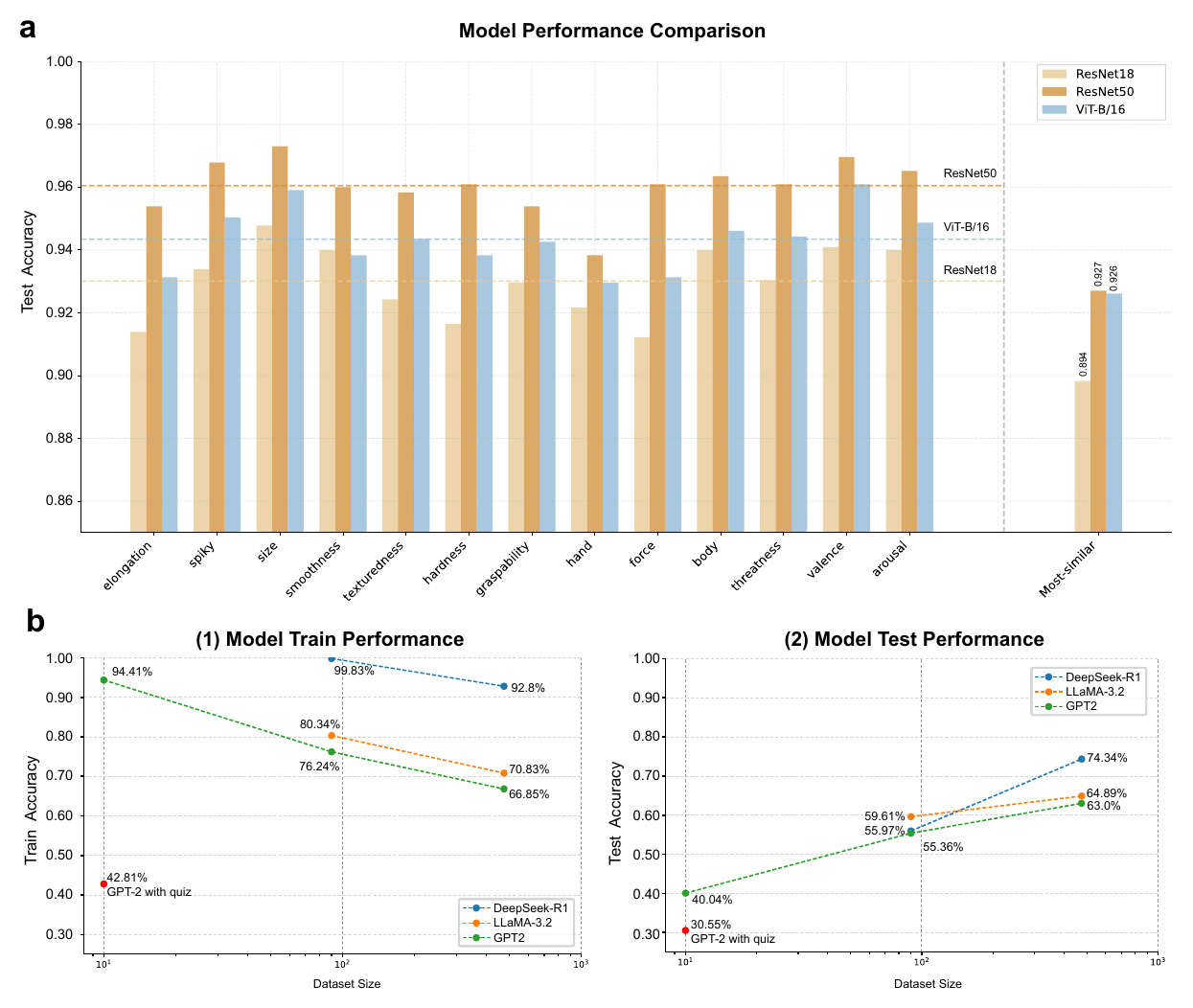}
    \caption{\textbf{Performance evaluation of visual and language models.} (a) Test accuracy of visual models (ResNet18, ResNet50, ViT-B/16) on attribute prediction and most similar class identification tasks. ResNet50 achieves the highest performance with 96.05\% attribute-wise accuracy and 92.70\% most similar classes accuracy. (b) Training and testing attribute-wise accuracy of language models (GPT-2, LLaMA-3.2-1.2B, DeepSeek-R1-1.5B) across different dataset sizes. The small dataset contains 10 training scenarios and 3 testing scenarios per tool, the medium dataset contains 90 training scenarios and 10 testing scenarios per tool, and the large dataset contains 475 training scenarios and 25 testing scenarios per tool. "GPT-2 with quiz" represents adding the question "What tool is relevant to this scene?" to scenario descriptions, which decreased performance compared to standard GPT-2. Larger models and datasets yield better performance, with DeepSeek-R1-1.5B demonstrates the best performance with 74.34\% attribute-wise accuracy on the largest dataset.}
    \label{fig:3}
\end{figure*}

\subsection{Visual Model Performance}
We evaluated three visual encoder architectures—ResNet18, ResNet50, and ViT-B/16—on their ability to predict tool attributes from images. As shown in Fig.\ref{fig:3}(a), all models demonstrated strong performance, with the ResNet50 architecture achieving the highest accuracy.

We evaluated performance using two complementary metrics. First, attribute-wise accuracy measures the model's ability to predict individual attribute values accurately on the 7-point scale. Specifically, the predicted values are rounded to the nearest integer on the 7-point scale, and the prediction is considered correct only if it exactly matches the ground truth value. ResNet50 achieved 96.05\% accuracy across all attributes and test samples, outperforming both ResNet18 (93.01\%) and ViT-B/16 (94.34\%). Second, most similar class accuracy evaluates whether the predicted attribute vector's closest matching tool category (by cosine similarity) matches the ground truth category. ResNet50 again demonstrated superior performance (92.70\%), closely followed by ViT-B/16 (92.61\%) and ResNet18 (89.40\%).

These results indicate that our visual pipeline effectively captures the physical and functional attributes of tools from images. The superior performance of ResNet50 suggests that medium-depth convolutional architectures strike an optimal balance between feature extraction capacity and generalization for attribute prediction. Notably, the high accuracy across all models demonstrates that visual feature extractors pretrained on general object recognition can be effectively repurposed for tool attribute prediction through targeted fine-tuning of prediction heads.

\begin{figure*}[t]
    \centering
    \includegraphics[width=1\linewidth]{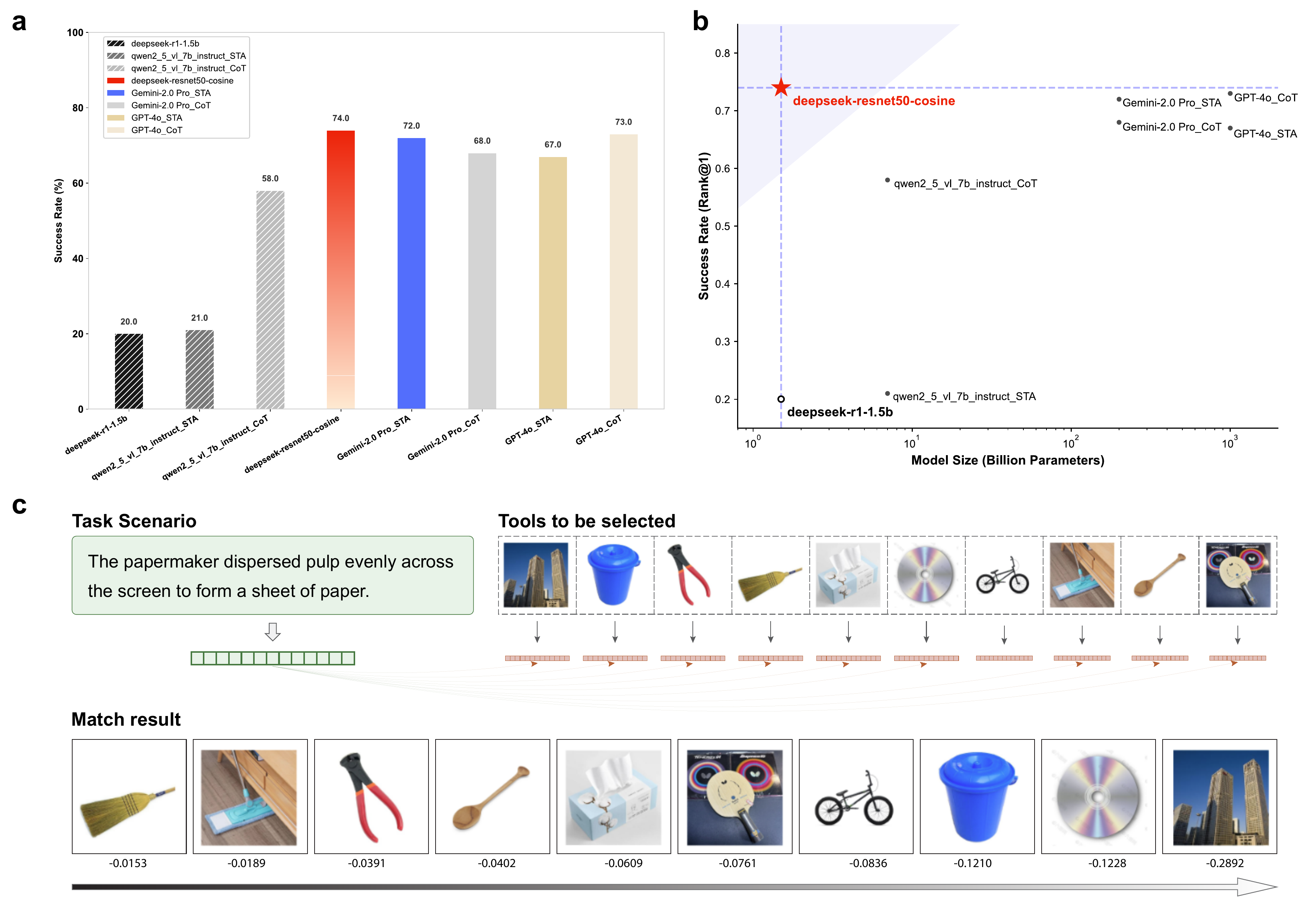}
    \caption{\textbf{Performance comparison of our attribute-based approach against baseline and larger multimodal models.} (a) Accuracy of different models on the tool selection task. Our approach using DeepSeek-R1-1.5B and ResNet50 achieved 74\% accuracy, significantly outperforming direct tool name matching (20\%) and smaller multimodal models. Performance comparison includes multimodal models with standard (STA) and chain-of-thought (CoT) prompting strategies: Qwen-VL-7B (STA: 21\%, CoT: 58\%), GPT-4o (STA: 67\%, CoT: 73\%), and Gemini-2.0-Pro (STA: 72\%, CoT: 68\%). (b) Model parameter efficiency comparison, showing our attribute-based approach (DeepSeek-R1-1.5B + ResNet50) achieves competitive performance with significantly fewer parameters compared to larger multimodal models. (c) Example visualization of tool ranking results from our attribute-based approach for a specific usage scenario, demonstrating the model's ability to correctly identify the most appropriate tool by matching scenario attributes with tool attributes.}
    \label{fig:4}
\end{figure*}

\subsection{Language Model Performance}
For the language pathway, we evaluated three progressively LLMs (GPT-2, LLaMA-3.2-1.2B and DeepSeek-R1-1.5B) for their ability to extract relevant attributes from textual task descriptions. We also investigated how the size of the dataset affects the performance of the model by training each model in three variants of the data set: small (10 training scenarios per tool), medium (90 scenarios per tool) and large (475 scenarios per tool).

As illustrated in Fig.\ref{fig:3}(b), we observed several key trends. First, in terms of the effect of model capacity, larger language models consistently outperformed smaller ones. DeepSeek-R1-1.5B achieved the highest attribute-wise accuracy (74.34\%), followed by LLaMA-3.2-1.2B (64.89\%) and GPT-2 (63.00\%) on the largest dataset. Second, considering the impact of the size of the dataset, all models benefited from increased training data, with performance improvements diminishing as the size of the dataset grew. Third, for all models, training accuracy actually decreased as dataset size increased, while testing accuracy consistently improved, indicating better generalization with more diverse training examples. DeepSeek-R1-1.5B demonstrated the best balance between high training performance and strong generalization.

Additionally, we tested whether explicitly adding a question prompt ("What tool is relevant to this scene?") at the end of each scenario description would improve attribute extraction. As shown in Fig.\ref{fig:3}(b), this modification surprisingly decreased both training accuracy (from 94.41\% to 42.81\%) and testing accuracy (from 63.00\% to 40.04\%) of GPT2 model, suggesting that explicit task framing may interfere with the model's ability to extract implicit attribute requirements.

These results highlight the challenge of extracting precise attribute requirements from natural language descriptions. Unlike visual attribute prediction, where accuracy exceeds 90\%, language-based attribute prediction remains more challenging, with the best model achieving 74.34\% accuracy. This performance gap likely reflects the inherent ambiguity in natural language descriptions and the implicit nature of attribute requirements in task descriptions.  Importantly, our results demonstrate a clear scaling trend: as both model capacity and dataset size increase, performance consistently improves.  GPT-2 (124M parameters) achieves 63.0\% accuracy, LLaMA-3.2-1.2B reaches 64.89\%, and DeepSeek-R1-1.5B attains 74.34\%. This scaling law suggests that further increases in model size and dataset expansion would likely yield additional performance gains.

\begin{figure*}[t]
    \centering
    \includegraphics[width=1\linewidth]{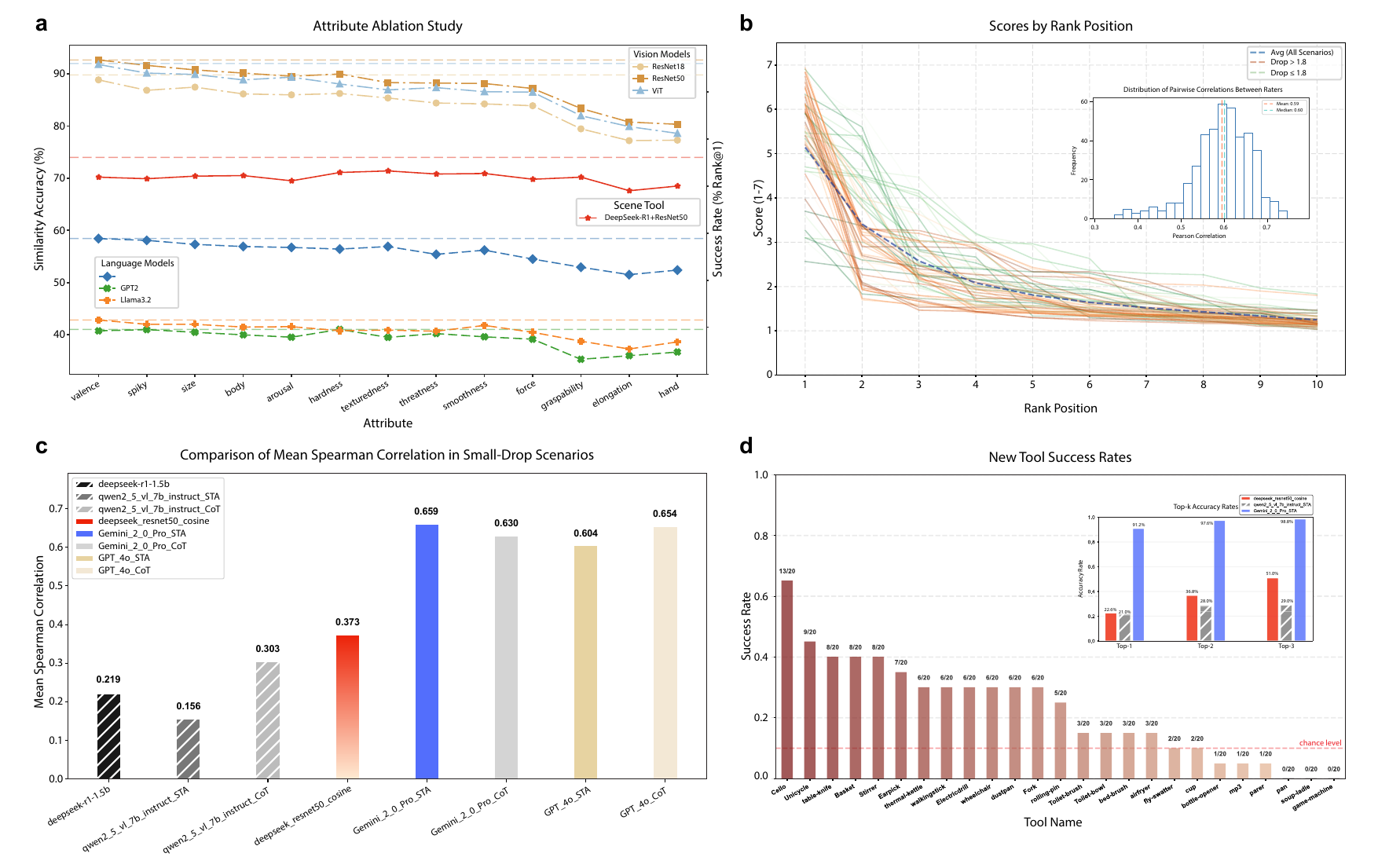}
    \caption{\textbf{Evaluation of attribute importance, human performance, and model generalization.} (a) Attribute ablation study showing the impact of removing individual attributes on performance across visual models, language models, and the combined system. Functional attributes (graspability, elongation, hand-relatedness) show the greatest impact. (b) Human performance evaluation (n=30) on 100 scenarios. Score distributions reveal scenarios with clear tool preferences (orange) versus comparable alternatives (green). Inset: distribution of inter-rater correlations (mean r = 0.59). (c) Comparison of mean Spearman correlation with human ratings across different models, analyzed specifically for scenarios with small preference gaps (separation $\leq 1.8$). Our deepseek\_resnet50\_cosine approach (r = 0.373) outperforms smaller models (qwen2\_5\_vl\_7b\_instruct
   and deepseek-r1-1.5b) but shows lower correlation than large multimodal models (Gemini\_2\_0\_Pro and GPT\_4o). (d) Generalization to 25 novel tools using our deepseek\_resnet50\_cosine approach shows variable success rates across individual tool categories. Inset: Top-k accuracy comparison across different models.}
    \label{fig:5}
\end{figure*}

\subsection{Tool Matching Performance}
To evaluate end-to-end performance on the Tool Matching Dataset, we compared our attribute-based approach against several baselines and state-of-the-art multimodal models. We first conducted an extensive evaluation of all possible language and vision model combinations to identify the optimal configuration for our framework. Table \ref{table2} presents the matching accuracy for all nine combinations of language models (GPT-2, LLaMA-3.2-1.2B, DeepSeek-R1-1.5B) and vision models (ResNet18, ResNet50, ViT-B/16). The results demonstrate clear performance scaling with model capacity and architectural choice. Among all combinations, DeepSeek-R1-1.5B paired with ResNet50 achieved the highest accuracy. Based on these findings, we selected the DeepSeek-R1-1.5B and ResNet50 combination for our comparative analysis with baseline methods and state-of-the-art multimodal models. As shown in Fig.\ref{fig:4}(a), our framework with this optimal configuration achieved 74\% accuracy on the tool selection task, where the system must select the correct tool from 10 candidates based on a textual scenario description.

\begin{table}[h]
\centering
\caption{Tool selection accuracy for different language-vision model combinations}
\begin{tabularx}{0.4\textwidth}{XXc}
\hline
\textbf{Language Model} & \textbf{Vision Model} & \textbf{Accuracy} \\
\hline
GPT2 & ResNet18 & 38\% \\
GPT2 & ViT-B/16 & 39\% \\
GPT2 & ResNet50 & 42\% \\
\hline
LLaMA-3.2-1.2B & ResNet18 & 57\% \\
LLaMA-3.2-1.2B & ViT-B/16 & 60\% \\
LLaMA-3.2-1.2B & ResNet50 & 62\% \\
\hline
DeepSeek-R1-1.5B & ResNet18 & 70\% \\
DeepSeek-R1-1.5B & ViT-B/16 & 72\% \\
DeepSeek-R1-1.5B & ResNet50 & 74\% \\
\hline
\end{tabularx}
\label{table2}
\end{table}

For comparison, we implemented several alternative methods. First, we established a direct naming baseline, a simple approach where we prompt the DeepSeek-R1-1.5B model to directly output the most appropriate tool name from a list of available tools based on the scenario description, without using an attribute-based intermediate representation. This approach achieved only 20\% accuracy, highlighting the limitations of direct mapping between scenario descriptions and tool names. Second, we tested Qwen-VL-7B, a smaller multimodal model with approximately 7 billion parameters. With straight-to-answer (STA) prompting, it achieved only 21\% accuracy, barely outperforming random selection. When enhanced with chain-of-thought (CoT) prompting, its performance improved significantly to 58\%, but still remained substantially below our attribute-based approach. Third, we evaluated two state-of-the-art large multimodal models: GPT-4o and Gemini-2.0-Pro. With standard prompting, GPT-4o achieved 67\% accuracy and Gemini-2.0-Pro achieved 72\%. With chain-of-thought prompting, GPT-4o improved to 73\%, while Gemini-2.0-Pro showed a slight decrease to 68\%.

As illustrated in Fig.\ref{fig:4}(b), our approach achieves competitive or superior performance compared to models with orders of magnitude more parameters. The combined parameter count of our DeepSeek-R1-1.5B language model and ResNet50 visual model is approximately 1.53 billion, compared to 7 billion for Qwen-VL-7B and estimated hundreds of billions for GPT-4o and Gemini-2.0-Pro. This remarkable efficiency stems from our approach's use of a structured attribute space that effectively bridges vision and language while requiring far fewer parameters than end-to-end multimodal training.

Fig.\ref{fig:4}(c) provides a qualitative example of our system's output for a specific scenario ("The papermaker dispersed pulp evenly across the screen to form a sheet of paper."). The visualization shows the ranking of candidate tools based on attribute similarity, with the broom correctly identified as the most suitable tool. The attribute vectors for both the scenario and candidate tools illustrate how the matching occurs in our shared attribute space.

\subsection{Ablation Studies}
To understand the relative importance of different attributes in our framework, we conducted an ablation study on an extended dataset of 1,000 tool selection scenarios, systematically removing individual attributes and measuring the impact on model performance. Fig. \ref{fig:5}(a) presents these results for visual encoders, language encoders, and the combined tool selection task.

For the visual encoder, consistent patterns emerged across all architectures. Hand-relatedness removal caused the most substantial performance degradation (12.35-13.39\%), followed closely by elongation (11.91-12.61\%) and graspability (9.30-10.35\%). In contrast, attributes like valence, spikiness, and size showed minimal impact when removed (generally below 3\% decrease). This suggests that functional characteristics related to human interaction and shape properties (particularly elongation) are the most informative visual cues for distinguishing between tool categories.

Similar patterns appeared in language models, where graspability, elongation, and hand-relatedness consistently proved most critical for accurate prediction. For DeepSeek-R1-1.5B, removing these functional attributes led to performance drops of 5-8\%, while psychological attributes like valence and arousal showed minimal impact. This consistency across modalities suggests that functional attributes related to tool manipulation and physical form are inherently more distinguishing in both visual and linguistic representations.

Consistent with the findings from the individual visual and language encoders, the ablation results for the end-to-end tool selection task reinforce the same attribute importance hierarchy. The removal of functional attributes resulted in the most significant performance degradation, while the exclusion of psychological attributes had a negligible impact. This reveals that our attribute space effectively captures the most salient properties for flexible tool selection, with functional and manipulation-related attributes proving particularly critical across modalities.

\subsection{Model-Human Alignment}
To evaluate our model's alignment with human decision-making, we conducted a evaluation with 30 human participants on the Tool Matching Dataset. Fig. \ref{fig:5}(b) presents the results of this human evaluation study. Human participants achieved 63\% accuracy in selecting the most appropriate tool from 10 candidates. The score distribution analysis reveals two distinct patterns: scenarios where participants showed comparable preferences for the top-ranked and second-ranked tools (separation $\leq 1.8$, shown in green), indicating multiple viable tool options, and scenarios with clear preference gaps (separation $
>1.8$, shown in orange), where one tool was decisively preferred over alternatives. Inter-rater reliability analysis showed moderate but significant agreement among human evaluators, with pairwise Pearson correlations averaging r = 0.59 ($p < 0.001$).

To assess model-human alignment, we calculated Spearman rank correlations between model predictions and averaged human ratings. Notably, the correlation analysis in Fig. 5(c) focuses specifically on scenarios with small preference gaps (separation $\leq 1.8$), where multiple tools were considered viable alternatives by human participants. Fig. \ref{fig:5}(c) shows that our attribute-based approach (deepseek\_resnet50\_cosine) achieved a mean correlation of r = 0.373 with human preferences. While this correlation is lower than large multimodal models like GPT-4o and Gemini-2.0-Pro, it substantially outperforms smaller baseline models, including the direct language approach (deepseek-r1-1.5b) and smaller multimodal models (qwen2\_5\_vl\_7b\_instruct). 

\subsection{Generalization to Novel Tools}
To evaluate our framework's ability to generalize beyond the training tool categories, we constructed a New Tool Matching Dataset featuring 25 completely new tool categories (Fig. S5). This evaluation tests whether our attribute-based approach can correctly identify appropriate tools for described scenarios even when encountering tool categories never seen during training.

Fig. \ref{fig:5}(d) presents the success rates for individual novel tools, revealing substantial variation in generalization performance. Some tools achieved success rates well above the 10\% chance level, with certain categories like cello and unicycle showing particularly strong performance. However, other novel tools, particularly those with unique functional properties or specialized applications, achieved success rates below chance level, indicating challenges in generalizing to tools with attribute combinations not well-represented in the training data.

Overall performance metrics demonstrate reasonable but limited generalization capability compared to large multimodal models. Our framework achieved 22.6\% Top-1 accuracy, 36.8\% Top-2 accuracy, and 51.0\% Top-3 accuracy on the novel tool dataset. In comparison, Qwen2.5-VL-7B-Instruct achieved 21.0\% Top-1 accuracy, 28.0\% Top-2 accuracy, and 29.0\% Top-3 accuracy, while Gemini-2.0-Pro demonstrated superior performance with 91.2\% Top-1 accuracy, 97.6\% Top-2 accuracy, and 98.8\% Top-3 accuracy. While our results exceed both chance performance (10\% for Top-1) and the smaller multimodal baseline, they indicate significant room for improvement in generalizing to completely unseen tool categories. The performance differences in novel tool generalization can be attributed primarily to training data exposure rather than architectural limitations. Our visual encoder was trained exclusively on the 115 tool categories in ToolNet, providing limited exposure to the diverse visual patterns present in novel tools. In contrast, large multimodal models like Gemini-2.0-Pro were trained on massive datasets containing millions of images across countless tool categories, enabling robust recognition of previously unseen tools.

These results highlight both the promise and limitations of our attribute-based approach for tool generalization, suggesting that while the framework can leverage learned attribute relationships to reason about novel tools, expanding the diversity of training tools and refining the attribute space could further improve generalization performance.

\section{Discussion and Conclusion}
\label{sec:discussion}

Our work establishes a cognitively inspired computational framework for flexible tool selection that achieves 74\% accuracy on standard benchmarks while using significantly fewer parameters than state-of-the-art multimodal models. Through evaluation across multiple dimensions—including model-human alignment analysis and novel tool generalization —we demonstrate both the computational efficiency and cognitive plausibility of attribute-based tool selection.

The performance gap between visual (96.05\%) and language (74.34\%) pathways reveals a fundamental asymmetry in how attributes are extracted versus inferred: visual systems can directly extract explicit physical properties from images, while language systems must infer implied requirements from natural language descriptions. This asymmetry aligns with cognitive research showing that physical object properties are more directly accessible than functional requirements inferred from task contexts\cite{dessalegn2013interaction,liao2024probing}. Our ablation studies provide crucial insights into the functional primitives underlying tool selection, revealing that manipulation-related attributes (graspability, hand-relatedness, elongation) consistently prove most critical across modalities. This consistency supports the technical reasoning hypothesis in cognitive science, which emphasizes the importance of functional and physical property reasoning in tool use\cite{renom2022exploring,bluet2025technical}. The minimal impact of psychological attributes (valence, arousal) suggests that affective properties, while potentially relevant in real-world contexts, are less critical for core tool selection decisions than functional characteristics.

The interpretability of our attribute space allows for systematic analysis of model decisions, addressing the opacity of large multimodal models\cite{liu2019tabby}. Our modular architecture enables independent optimization of visual and language components, facilitating iterative refinement and adaptation across domains. The parameter efficiency of this approach makes it particularly suitable for resource-constrained applications where computational overhead is critical. Moreover, our attribute-based architecture provides a testable computational model for neurocognitive research\cite{loosen2025revisiting}, potentially informing psychophysical experiments investigating human attribute prioritization and neuroimaging studies examining neural correlates of different tool properties.

Our novel tool generalization experiment reveals both the promise and limitations of attribute-based approaches for handling unseen tool categories. The 22.6\% Top-1 accuracy on completely novel tools, while exceeding chance performance (10\%), indicates that learned attribute relationships enable some transfer to new categories. However, the substantial variation across individual novel tools (ranging from well above chance to below chance performance) suggests that generalization success depends heavily on the similarity between novel tool attributes and those represented in the training dataset.


While our framework demonstrates significant advances, several limitations warrant consideration and suggest directions for future work. First, despite being theoretically motivated and empirically validated, our 13-dimensional attribute space represents a simplification of human tool cognition. These dimensions capture essential physical and functional properties but may not encompass all factors relevant to real-world tool selection, such as availability, cost, or contextual appropriateness\cite{baber2014tool}. Second, the visual-language performance gap suggests room for improvement in extracting attribute requirements from natural language descriptions. Third, our evaluation focuses on static tool selection rather than dynamic manipulation\cite{choi2021use}, which would require additional considerations of temporal sequences and motor control. Fourth, the moderate model-human correlation (r = 0.373) suggests our attribute space may miss important factors in human decision-making, such as tool availability, safety concerns, or personal familiarity. Fifth, our generalization experiment reveals a fundamental limitation: while maintaining competitive performance with large multimodal models on known tools (74\% vs 72-73\%), our approach shows substantial performance drops on novel tool categories (22.6\% vs 91.2\% Top-1 accuracy for Gemini-2.0-Pro). This disparity stems primarily from training data limitations rather than architectural constraints—our visual encoder trained on 115 tool categories cannot match the recognition capabilities of models trained on millions of diverse tool images. Future research should explore expanding the attribute space to incorporate additional dimensions (e.g., material properties, temporal constraints) while investigating hierarchical or learned attribute spaces through unsupervised methods to discover optimal representations for specific domains. Extending the framework to dynamic tool manipulation and sequential task planning would better capture the full complexity of human tool use and its neural mechanisms. 

This work advances our understanding of flexible tool selection through a computationally efficient and cognitively plausible framework that bridges visual perception and linguistic understanding. By demonstrating that attribute-based representations enable effective cross-modal matching for tool selection—while revealing important limitations in human alignment and novel tool generalization—we contribute to both cognitive science and computational modeling of human-like intelligent systems. Our evaluation across standard performance, human comparison, and generalization scenarios provides a foundation for developing more interpretable, efficient, and neurally-grounded systems that reflect both the capabilities and limitations of human tool use. Future advances in attribute space design and generalization methods will be crucial for deploying such systems in real-world applications where tool diversity and contextual complexity exceed current capabilities.

\section*{Code Availability}
The complete source code for our attribute-based tool selection framework and the datasets are available at: \url{https://drive.google.com/drive/folders/179W5KxPMzidjgg4hTzObrz-RvnXcVW0q?usp=sharing}. Pre-trained model weights for all configurations tested in this paper are available at: \url{https://1024terabox.com/s/17IdOOZ12GVtwJiWKySkOeQ}




\bibliographystyle{IEEEtran}
\bibliography{IEEEabrv,ref}

\begin{thebibliography}{10}
\providecommand{\url}[1]{#1}
\csname url@samestyle\endcsname
\providecommand{\newblock}{\relax}
\providecommand{\bibinfo}[2]{#2}
\providecommand{\BIBentrySTDinterwordspacing}{\spaceskip=0pt\relax}
\providecommand{\BIBentryALTinterwordstretchfactor}{4}
\providecommand{\BIBentryALTinterwordspacing}{\spaceskip=\fontdimen2\font plus
\BIBentryALTinterwordstretchfactor\fontdimen3\font minus \fontdimen4\font\relax}
\providecommand{\BIBforeignlanguage}[2]{{%
\expandafter\ifx\csname l@#1\endcsname\relax
\typeout{** WARNING: IEEEtran.bst: No hyphenation pattern has been}%
\typeout{** loaded for the language `#1'. Using the pattern for}%
\typeout{** the default language instead.}%
\else
\language=\csname l@#1\endcsname
\fi
#2}}
\providecommand{\BIBdecl}{\relax}
\BIBdecl

\bibitem{baber_2003_cognition}
C.~Baber, \emph{Cognition and tool use: Forms of engagement in human and animal use of tools}.\hskip 1em plus 0.5em minus 0.4em\relax CRC Press, 2003.

\bibitem{johnsonfrey_2003_whats}
S.~H. Johnson-Frey, ``What's so special about human tool use?'' \emph{Neuron}, vol.~39, pp. 201--204, 07 2003.

\bibitem{goodall1964tool}
J.~Goodall, ``Tool-using and aimed throwing in a community of free-living chimpanzees,'' \emph{Nature}, vol. 201, no. 4926, pp. 1264--1266, 1964.

\bibitem{thorpe2009orangutans}
S.~K. Thorpe, R.~Holder, and R.~H. Crompton, ``Orangutans employ unique strategies to control branch flexibility,'' \emph{Proceedings of the National Academy of Sciences}, vol. 106, no.~31, pp. 12\,646--12\,651, 2009.

\bibitem{fellers1976tool}
J.~H. Fellers and G.~M. Fellers, ``Tool use in a social insect and its implications for competitive interactions,'' \emph{Science}, vol. 192, no. 4234, pp. 70--72, 1976.

\bibitem{stout2011stone}
D.~Stout, ``Stone toolmaking and the evolution of human culture and cognition,'' \emph{Philosophical Transactions of the Royal Society B: Biological Sciences}, vol. 366, no. 1567, pp. 1050--1059, 2011.

\bibitem{morgan2015experimental}
T.~J. Morgan, N.~T. Uomini, L.~E. Rendell, L.~Chouinard-Thuly, S.~E. Street, H.~M. Lewis, C.~P. Cross, C.~Evans, R.~Kearney, I.~de~la Torre \emph{et~al.}, ``Experimental evidence for the co-evolution of hominin tool-making teaching and language,'' \emph{Nature communications}, vol.~6, no.~1, p. 6029, 2015.

\bibitem{biro2013tool}
D.~Biro, M.~Haslam, and C.~Rutz, ``Tool use as adaptation,'' p. 20120408, 2013.

\bibitem{heersmink2022human}
R.~Heersmink, ``Human uniqueness in using tools and artifacts: flexibility, variety, complexity,'' \emph{Synthese}, vol. 200, no.~6, p. 442, 2022.

\bibitem{dehaene2022symbols}
S.~Dehaene, F.~Al~Roumi, Y.~Lakretz, S.~Planton, and M.~Sabl{\'e}-Meyer, ``Symbols and mental programs: a hypothesis about human singularity,'' \emph{Trends in Cognitive Sciences}, vol.~26, no.~9, pp. 751--766, 2022.

\bibitem{johnson2004neural}
S.~H. Johnson-Frey, ``The neural bases of complex tool use in humans,'' \emph{Trends in cognitive sciences}, vol.~8, no.~2, pp. 71--78, 2004.

\bibitem{goldenberg_2009_the}
G.~Goldenberg and J.~Spatt, ``The neural basis of tool use,'' \emph{Brain}, vol. 132, pp. 1645--1655, 04 2009.

\bibitem{vaesen_2012_the}
K.~Vaesen, ``The cognitive bases of human tool use,'' \emph{Behavioral and Brain Sciences}, vol.~35, pp. 203--218, 06 2012.

\bibitem{federico2023functional}
G.~Federico, F.~Osiurak, G.~Ciccarelli, C.~R. Ilardi, C.~Cavaliere, L.~Tramontano, V.~Alfano, M.~Migliaccio, A.~Di~Cecca, M.~Salvatore \emph{et~al.}, ``On the functional brain networks involved in tool-related action understanding,'' \emph{Communications Biology}, vol.~6, no.~1, p. 1163, 2023.

\bibitem{martin1996neural}
A.~Martin, C.~L. Wiggs, L.~G. Ungerleider, and J.~V. Haxby, ``Neural correlates of category-specific knowledge,'' \emph{Nature}, vol. 379, no. 6566, pp. 649--652, 1996.

\bibitem{kellenbach2003actions}
M.~L. Kellenbach, M.~Brett, and K.~Patterson, ``Actions speak louder than functions: the importance of manipulability and action in tool representation,'' \emph{Journal of cognitive neuroscience}, vol.~15, no.~1, pp. 30--46, 2003.

\bibitem{huth2012continuous}
A.~G. Huth, S.~Nishimoto, A.~T. Vu, and J.~L. Gallant, ``A continuous semantic space describes the representation of thousands of object and action categories across the human brain,'' \emph{Neuron}, vol.~76, no.~6, pp. 1210--1224, 2012.

\bibitem{osiurak_2018_looking}
F.~Osiurak and D.~Heinke, ``Looking for intoolligence: A unified framework for the cognitive study of human tool use and technology.'' \emph{American Psychologist}, vol.~73, pp. 169--185, 02 2018.

\bibitem{Saito_2021_How}
N.~Saito, T.~Ogata, S.~Funabashi, H.~Mori, and S.~Sugano, ``How to select and use tools? : Active perception of target objects using multimodal deep learning,'' \emph{IEEE Robotics and Automation Letters}, vol.~6, no.~2, pp. 2517--2524, 2021.

\bibitem{fischer2021tool}
J.~Fischer and B.~Z. Mahon, ``What tool representation, intuitive physics, and action have in common: The brain’s first-person physics engine,'' \emph{Cognitive neuropsychology}, vol.~38, no. 7-8, pp. 455--467, 2021.

\bibitem{hurst2024gpt}
A.~Hurst, A.~Lerer, A.~P. Goucher, A.~Perelman, A.~Ramesh, A.~Clark, A.~Ostrow, A.~Welihinda, A.~Hayes, A.~Radford \emph{et~al.}, ``Gpt-4o system card,'' \emph{arXiv preprint arXiv:2410.21276}, 2024.

\bibitem{gemini20pro}
\BIBentryALTinterwordspacing
Google, ``Gemini 2.0 pro model card,'' Google Cloud Platform, Vertex AI, February 2025, experimental Model. [Online]. Available: \url{https://www.prompthub.us/models/gemini-2-0-pro}
\BIBentrySTDinterwordspacing

\bibitem{peeters2009representation}
R.~Peeters, L.~Simone, K.~Nelissen, M.~Fabbri-Destro, W.~Vanduffel, G.~Rizzolatti, and G.~A. Orban, ``The representation of tool use in humans and monkeys: common and uniquely human features,'' \emph{Journal of Neuroscience}, vol.~29, no.~37, pp. 11\,523--11\,539, 2009.

\bibitem{gallivan2013decoding}
J.~P. Gallivan, D.~A. McLean, K.~F. Valyear, and J.~C. Culham, ``Decoding the neural mechanisms of human tool use,'' \emph{elife}, vol.~2, p. e00425, 2013.

\bibitem{orban2014neural}
G.~A. Orban and F.~Caruana, ``The neural basis of human tool use,'' \emph{Frontiers in psychology}, vol.~5, p. 310, 2014.

\bibitem{maravita2018parietal}
A.~Maravita and D.~Romano, ``The parietal lobe and tool use,'' \emph{Handbook of clinical neurology}, vol. 151, pp. 481--498, 2018.

\bibitem{grafton1997premotor}
S.~T. Grafton, L.~Fadiga, M.~A. Arbib, and G.~Rizzolatti, ``Premotor cortex activation during observation and naming of familiar tools,'' \emph{Neuroimage}, vol.~6, no.~4, pp. 231--236, 1997.

\bibitem{cabrera2020neural}
M.~J. Cabrera-{\'A}lvarez and N.~S. Clayton, ``Neural processes underlying tool use in humans, macaques, and corvids,'' \emph{Frontiers in Psychology}, vol.~11, p. 560669, 2020.

\bibitem{lesourd2021semantic}
M.~Lesourd, M.~Servant, J.~Baumard, E.~Reynaud, C.~Ecochard, F.~T. Medjaoui, A.~Bartolo, and F.~Osiurak, ``Semantic and action tool knowledge in the brain: Identifying common and distinct networks,'' \emph{Neuropsychologia}, vol. 159, p. 107918, 2021.

\bibitem{cortinovis2025tool}
D.~Cortinovis, M.~V. Peelen, and S.~Bracci, ``Tool representations in human visual cortex,'' \emph{Journal of Cognitive Neuroscience}, vol.~37, no.~3, pp. 515--531, 2025.

\bibitem{mangalam2022psychological}
M.~Mangalam, D.~M. Fragaszy, J.~B. Wagman, B.~M. Day, D.~G. Kelty-Stephen, R.~M. Bongers, D.~W. Stout, and F.~Osiurak, ``On the psychological origins of tool use,'' \emph{Neuroscience \& Biobehavioral Reviews}, vol. 134, p. 104521, 2022.

\bibitem{osiurak2010grasping}
F.~Osiurak, C.~Jarry, and D.~Le~Gall, ``Grasping the affordances, understanding the reasoning: toward a dialectical theory of human tool use.'' \emph{Psychological review}, vol. 117, no.~2, p. 517, 2010.

\bibitem{osiurak2016tool}
F.~Osiurak and A.~Badets, ``Tool use and affordance: Manipulation-based versus reasoning-based approaches.'' \emph{Psychological review}, vol. 123, no.~5, p. 534, 2016.

\bibitem{goldenberg1998tool}
G.~Goldenberg and S.~Hagmann, ``Tool use and mechanical problem solving in apraxia,'' \emph{Neuropsychologia}, vol.~36, no.~7, pp. 581--589, 1998.

\bibitem{osiurak2009unusual}
F.~Osiurak, C.~Jarry, P.~Allain, G.~Aubin, F.~Etcharry-Bouyx, I.~Richard, I.~Bernard, and D.~Le~Gall, ``Unusual use of objects after unilateral brain damage. the technical reasoning model,'' \emph{Cortex}, vol.~45, no.~6, pp. 769--783, 2009.

\bibitem{buxbaum2017learning}
L.~J. Buxbaum, ``Learning, remembering, and predicting how to use tools: Distributed neurocognitive mechanisms: Comment on osiurak and badets (2016).'' \emph{Psychological Review}, 2017.

\bibitem{allen2020rapid}
K.~R. Allen, K.~A. Smith, and J.~B. Tenenbaum, ``Rapid trial-and-error learning with simulation supports flexible tool use and physical reasoning,'' \emph{Proceedings of the National Academy of Sciences}, vol. 117, no.~47, pp. 29\,302--29\,310, 2020.

\bibitem{antol2015vqa}
S.~Antol, A.~Agrawal, J.~Lu, M.~Mitchell, D.~Batra, C.~L. Zitnick, and D.~Parikh, ``Vqa: Visual question answering,'' in \emph{Proceedings of the IEEE international conference on computer vision}, 2015, pp. 2425--2433.

\bibitem{yang2016stacked}
Z.~Yang, X.~He, J.~Gao, L.~Deng, and A.~Smola, ``Stacked attention networks for image question answering,'' in \emph{Proceedings of the IEEE conference on computer vision and pattern recognition}, 2016, pp. 21--29.

\bibitem{anderson2018bottom}
P.~Anderson, X.~He, C.~Buehler, D.~Teney, M.~Johnson, S.~Gould, and L.~Zhang, ``Bottom-up and top-down attention for image captioning and visual question answering,'' in \emph{Proceedings of the IEEE conference on computer vision and pattern recognition}, 2018, pp. 6077--6086.

\bibitem{lu2019vilbert}
J.~Lu, D.~Batra, D.~Parikh, and S.~Lee, ``Vilbert: Pretraining task-agnostic visiolinguistic representations for vision-and-language tasks,'' \emph{Advances in neural information processing systems}, vol.~32, 2019.

\bibitem{chen2020uniter}
Y.-C. Chen, L.~Li, L.~Yu, A.~El~Kholy, F.~Ahmed, Z.~Gan, Y.~Cheng, and J.~Liu, ``Uniter: Universal image-text representation learning,'' in \emph{European conference on computer vision}.\hskip 1em plus 0.5em minus 0.4em\relax Springer, 2020, pp. 104--120.

\bibitem{radford2021learning}
A.~Radford, J.~W. Kim, C.~Hallacy, A.~Ramesh, G.~Goh, S.~Agarwal, G.~Sastry, A.~Askell, P.~Mishkin, J.~Clark \emph{et~al.}, ``Learning transferable visual models from natural language supervision,'' in \emph{International conference on machine learning}.\hskip 1em plus 0.5em minus 0.4em\relax PmLR, 2021, pp. 8748--8763.

\bibitem{qian2024linguistic}
S.~Qian, Z.~Zhou, D.~Xue, B.~Wang, and C.~Xu, ``From linguistic giants to sensory maestros: A survey on cross-modal reasoning with large language models,'' \emph{arXiv preprint arXiv:2409.18996}, 2024.

\bibitem{xu2024lvlm}
P.~Xu, W.~Shao, K.~Zhang, P.~Gao, S.~Liu, M.~Lei, F.~Meng, S.~Huang, Y.~Qiao, and P.~Luo, ``Lvlm-ehub: A comprehensive evaluation benchmark for large vision-language models,'' \emph{IEEE Transactions on Pattern Analysis and Machine Intelligence}, 2024.

\bibitem{zhang2024scaling}
B.~Zhang, Z.~Liu, C.~Cherry, and O.~Firat, ``When scaling meets llm finetuning: The effect of data, model and finetuning method,'' in \emph{ICLR}, 2024.

\bibitem{bilal2025llms_tist}
A.~Bilal, D.~Ebert, and B.~Lin, ``{LLMs for Explainable AI: A Comprehensive Survey},'' \emph{ACM Transactions on Intelligent Systems and Technology}, mar 2025.

\bibitem{xie2019improvisation}
A.~Xie, F.~Ebert, S.~Levine, and C.~Finn, ``Improvisation through physical understanding: Using novel objects as tools with visual foresight,'' \emph{arXiv preprint arXiv:1904.05538}, 2019.

\bibitem{myers2015affordance}
A.~Myers, C.~L. Teo, C.~Ferm{\"u}ller, and Y.~Aloimonos, ``Affordance detection of tool parts from geometric features,'' in \emph{2015 IEEE international conference on robotics and automation (ICRA)}.\hskip 1em plus 0.5em minus 0.4em\relax IEEE, 2015, pp. 1374--1381.

\bibitem{huang2024metatool}
Y.~Huang, J.~Shi, Y.~Li, C.~Fan, S.~Wu, Q.~Zhang, Y.~Liu, P.~Zhou, Y.~Wan, N.~Z. Gong \emph{et~al.}, ``Metatool benchmark for large language models: Deciding whether to use tools and which to use,'' in \emph{ICLR}, 2024.

\bibitem{sharif2014cnn}
A.~Sharif~Razavian, H.~Azizpour, J.~Sullivan, and S.~Carlsson, ``Cnn features off-the-shelf: an astounding baseline for recognition,'' in \emph{Proceedings of the IEEE conference on computer vision and pattern recognition workshops}, 2014, pp. 806--813.

\bibitem{kornblith2019better}
S.~Kornblith, J.~Shlens, and Q.~V. Le, ``Do better imagenet models transfer better?'' in \emph{Proceedings of the IEEE/CVF conference on computer vision and pattern recognition}, 2019, pp. 2661--2671.

\bibitem{he2016deep}
K.~He, X.~Zhang, S.~Ren, and J.~Sun, ``Deep residual learning for image recognition,'' in \emph{Proceedings of the IEEE conference on computer vision and pattern recognition}, 2016, pp. 770--778.

\bibitem{dosovitskiy2020image}
A.~Dosovitskiy, L.~Beyer, A.~Kolesnikov, D.~Weissenborn, X.~Zhai, T.~Unterthiner, M.~Dehghani, M.~Minderer, G.~Heigold, S.~Gelly, J.~Uszkoreit, and N.~Houlsby, ``An image is worth 16x16 words: Transformers for image recognition at scale,'' in \emph{ICLR}, 2020.

\bibitem{ninama2024computer}
H.~Ninama, J.~Raikwal, A.~Ravuri, D.~Sukheja, S.~K. Bhoi, N.~Jhanjhi, A.~A.~H. Elnour, and A.~Abdelmaboud, ``Computer vision and deep transfer learning for automatic gauge reading detection,'' \emph{Scientific Reports}, vol.~14, no.~1, p. 23019, 2024.

\bibitem{zeng2019continual}
G.~Zeng, Y.~Chen, B.~Cui, and S.~Yu, ``Continual learning of context-dependent processing in neural networks,'' \emph{Nature Machine Intelligence}, vol.~1, no.~8, pp. 364--372, 2019.

\bibitem{davila2024comparison}
A.~Davila, J.~Colan, and Y.~Hasegawa, ``Comparison of fine-tuning strategies for transfer learning in medical image classification,'' \emph{Image and Vision Computing}, vol. 146, p. 105012, 2024.

\bibitem{rebuffi2017learning}
S.-A. Rebuffi, H.~Bilen, and A.~Vedaldi, ``Learning multiple visual domains with residual adapters,'' \emph{Advances in neural information processing systems}, vol.~30, 2017.

\bibitem{beyer2022knowledge}
L.~Beyer, X.~Zhai, A.~Royer, L.~Markeeva, R.~Anil, and A.~Kolesnikov, ``Knowledge distillation: A good teacher is patient and consistent,'' in \emph{Proceedings of the IEEE/CVF conference on computer vision and pattern recognition}, 2022, pp. 10\,925--10\,934.

\bibitem{vaswani2017attention}
A.~Vaswani, N.~Shazeer, N.~Parmar, J.~Uszkoreit, L.~Jones, A.~N. Gomez, {\L}.~Kaiser, and I.~Polosukhin, ``Attention is all you need,'' \emph{Advances in neural information processing systems}, vol.~30, 2017.

\bibitem{radford2018improving}
A.~Radford, K.~Narasimhan, T.~Salimans, I.~Sutskever \emph{et~al.}, ``Improving language understanding by generative pre-training,'' 2018.

\bibitem{devlin2019bert}
J.~Devlin, M.-W. Chang, K.~Lee, and K.~Toutanova, ``Bert: Pre-training of deep bidirectional transformers for language understanding,'' in \emph{Proceedings of the 2019 conference of the North American chapter of the association for computational linguistics: human language technologies, volume 1 (long and short papers)}, 2019, pp. 4171--4186.

\bibitem{touvron2023llama}
H.~Touvron, L.~Martin, K.~Stone, P.~Albert, A.~Almahairi, Y.~Babaei, N.~Bashlykov, S.~Batra, P.~Bhargava, S.~Bhosale \emph{et~al.}, ``Llama 2: Open foundation and fine-tuned chat models,'' \emph{arXiv preprint arXiv:2307.09288}, 2023.

\bibitem{liu2024deepseek}
A.~Liu, B.~Feng, B.~Xue, B.~Wang, B.~Wu, C.~Lu, C.~Zhao, C.~Deng, C.~Zhang, C.~Ruan \emph{et~al.}, ``Deepseek-v3 technical report,'' \emph{arXiv preprint arXiv:2412.19437}, 2024.

\bibitem{zhou2020evaluating}
X.~Zhou, Y.~Zhang, L.~Cui, and D.~Huang, ``Evaluating commonsense in pre-trained language models,'' in \emph{Proceedings of the AAAI conference on artificial intelligence}, vol.~34, no.~05, 2020, pp. 9733--9740.

\bibitem{howard2018universal}
J.~Howard and S.~Ruder, ``Universal language model fine-tuning for text classification,'' \emph{arXiv preprint arXiv:1801.06146}, 2018.

\bibitem{lester2021power}
B.~Lester, R.~Al-Rfou, and N.~Constant, ``The power of scale for parameter-efficient prompt tuning,'' \emph{arXiv preprint arXiv:2104.08691}, 2021.

\bibitem{li2021prefix}
X.~L. Li and P.~Liang, ``Prefix-tuning: Optimizing continuous prompts for generation,'' \emph{arXiv preprint arXiv:2101.00190}, 2021.

\bibitem{hu2022lora}
E.~J. Hu, Y.~Shen, P.~Wallis, Z.~Allen-Zhu, Y.~Li, S.~Wang, L.~Wang, W.~Chen \emph{et~al.}, ``Lora: Low-rank adaptation of large language models.'' \emph{ICLR}, vol.~1, no.~2, p.~3, 2022.

\bibitem{dessalegn2013interaction}
B.~Dessalegn and B.~Landau, ``Interaction between language and vision: It’s momentary, abstract, and it develops,'' \emph{Cognition}, vol. 127, no.~3, pp. 331--344, 2013.

\bibitem{liao2024probing}
C.~Liao, M.~Sawayama, and B.~Xiao, ``Probing the link between vision and language in material perception using psychophysics and unsupervised learning,'' \emph{PLOS Computational Biology}, vol.~20, no.~10, p. e1012481, 2024.

\bibitem{renom2022exploring}
M.~A. Renom, B.~Caramiaux, and M.~Beaudouin-Lafon, ``Exploring technical reasoning in digital tool use,'' in \emph{Proceedings of the 2022 CHI Conference on Human Factors in Computing Systems}, 2022, pp. 1--17.

\bibitem{bluet2025technical}
A.~Bluet, E.~Reynaud, G.~Federico, C.~Bryche, M.~Lesourd, A.~Fournel, F.~Lamberton, D.~Ibarrola, Y.~Rossetti, and F.~Osiurak, ``The technical-reasoning network is recruited when people observe others make or teach how to make tools: An fmri study,'' \emph{iScience}, vol.~28, no.~2, 2025.

\bibitem{liu2019tabby}
H.~Liu, R.~Wang, S.~Shan, and X.~Chen, ``What is a tabby? interpretable model decisions by learning attribute-based classification criteria,'' \emph{IEEE transactions on pattern analysis and machine intelligence}, vol.~43, no.~5, pp. 1791--1807, 2019.

\bibitem{loosen2025revisiting}
A.~M. Loosen, A.~Kato, and X.~Gu, ``Revisiting the role of computational neuroimaging in the era of integrative neuroscience,'' \emph{Neuropsychopharmacology}, vol.~50, no.~1, pp. 103--113, 2025.

\bibitem{baber2014tool}
C.~Baber, M.~Parekh, and T.~G. Cengiz, ``Tool use as distributed cognition: how tools help, hinder and define manual skill,'' \emph{Frontiers in psychology}, vol.~5, p. 116, 2014.

\bibitem{choi2021use}
H.~Choi, C.~Crump, C.~Duriez, A.~Elmquist, G.~Hager, D.~Han, F.~Hearl, J.~Hodgins, A.~Jain, F.~Leve \emph{et~al.}, ``On the use of simulation in robotics: Opportunities, challenges, and suggestions for moving forward,'' \emph{Proceedings of the National Academy of Sciences}, vol. 118, no.~1, p. e1907856118, 2021.

\end{thebibliography}

\vfill

\end{document}